\documentclass[journal]{IEEEtran}

\usepackage{url}
\usepackage{amsmath}
\usepackage{amsfonts}
\usepackage{graphicx}
\usepackage{subfig}
\usepackage{multirow}

\begin{document}

\title{Prediction of Sea Surface Temperature using Long Short-Term Memory}

\author{Qin~Zhang,
        ~Hui~Wang,
        ~Junyu~Dong,~\IEEEmembership{Member,~IEEE}
        ~Guoqiang~Zhong,~\IEEEmembership{Member,~IEEE}
        and~Xin~Sun~\IEEEmembership{Member,~IEEE}
\thanks{This work was supported in part by the National Natural Science Foundation of China (No.41576011, 61403353, 61401413) and International Science \& Technology Cooperation Program of China(ISTCP)(N0.2014DFA10410)}
\thanks{Q. Zhang, J. Dong, G. Zhong, and X. Sun are with the Department of Computer Science and Technology,
Ocean University of China, Qingdao 266100, China (corresponding e-mail: dongjunyu@ouc.edu.cn).}
\thanks{Q. Zhang is also with the Department of Science and Information, Agriculture University of Qingdao, Qingdao 266109, China.}
\thanks{H. Wang is with the College of Oceanic and Atmospheric Sciences,
Ocean University of China, Qingdao 266100, China.}
}

\markboth{SUBMITTED TO IEEE GEOSCIENCE AND REMOTE SENSING LETTERS}%
{Shell }

\maketitle

\begin{abstract}
  This letter adopts long short-term memory(LSTM) to predict sea surface temperature(SST), which is the first attempt, to our knowledge, to use recurrent neural network to solve the problem of SST prediction, and to make one week and one month daily prediction. We formulate the SST prediction problem as a time series regression problem. LSTM is a special kind of recurrent neural network, which introduces gate mechanism into vanilla RNN to prevent the vanished or exploding gradient problem. It has strong ability to model the temporal relationship of time series data and can handle the long-term dependency problem well. The proposed network architecture is composed of two kinds of layers: LSTM layer and full-connected dense layer. LSTM layer is utilized to model the time series relationship. Full-connected layer is utilized to map the output of LSTM layer to a final prediction. We explore the optimal setting of this architecture by experiments and report the accuracy of coastal seas of China to confirm the effectiveness of the proposed method. In addition, we also show its online updated characteristics.
\end{abstract}


\IEEEpeerreviewmaketitle

\section{Introduction}

\IEEEPARstart{S}{ea} surface temperature, SST for short, is an important parameter in the energy balance system of the earth's surface, and it is also a critical indicator to measure the sea water heat. It plays an important role in the process of the earth's surface and atmosphere interaction. Sea occupies three quarters of the global area, therefore SST has inestimable influence on the global climate and the biological systems. In recent years, people focus more and more on sea surface temperature. The prediction of SST becomes a hot research increasingly. It is also an important and fundamental problem in many application domains such as forecasting ocean weather and climate, offshore activities like fishing and mining, ocean environment protection, ocean military affairs, etc. It is significant in science research and application to predict accurate temporal and spatial distribution for SST. However, its prediction accuracy is always low because of many uncertain factors especially in coastal seas.

Many methods have been published to predict the sea surface temperature. These methods can be generally classified into two categories according to the different point of view to create models~\cite{Patil2016Prediction}. One is based on physics, also known as numerical model. The other is based on data, also called data-driven model. The former tries to utilize a series of differential equations to describe the variation of SST, which is usually sophisticated and demands increasing computational effort and time. In addition, numerical model differs in different sea areas. While the latter tries to learn the model from data. Some learning methods were used such as linear regression~\cite{Kug2004A}, Support Vector Machines~\cite{Lins2010Sea}, Neural Network~\cite{Patil2016Prediction}, etc.

This letter focuses on the second way to predict SST, which uses long short-term memory (LSTM) to model time series of SST data. Long short-term memory is a special kind of recurrent neural network (RNN), which is a class of artificial neural network where connections between units form a directed cycle. This creates an internal state of the network which allows it to exhibit dynamic temporal behavior. Unlike feedforward neural networks, RNNs can use their internal memory to process arbitrary sequences of inputs~\cite{rnnwebpage}. However, vanilla RNN is difficult to train and suffers a lot about vanishing or exploding gradient problem, which can not solve the long-term dependency problem. While LSTM introduces gate mechanism to prevent backpropagated errors from vanishing or exploding, which has been subsequently proved to be more effective than conventional RNNs~\cite{Lecun2015Deep}.

In this letter, a LSTM based prediction method for SST is proposed. Our main contributions are twofold: 1) a LSTM based network is properly designed with full connected layer to form a regression model for SST prediction. LSTM layer is utilized to catch the temporal relationship among SST time series data. Full connected layer is utilized to map the output of LSTM layer to a final prediction. 2) SST changes relatively stable in ocean, while more fluctuated in coastal seas. We focus on the latter, and report the prediction accuracy beyond the existing methods, which confirms the effectiveness of the proposed method.

The remainder of this letter is organized as follows. Section II gives the problem formulation and describes the proposed method in detail. Experimental results on Bohai SST Dataset, which is chosen from NOAA OI SST V2 High Resolution Dataset are reported in Section III. Finally, Section IV concludes this letter.

\section{Methodology}

\subsection{Problem formulation}

Usually according to the latitude and longitude the sea surface can be divided into grids. Each grid will have a value at every time interval. Then the SST values can be organized as three dimensional grids. The problem is how to predict the future value of SST given this 3D SST grid.

Suppose we take the SST values from one single grid along all the time. It is a time series real values. If we can build a model to capture the temporal relationship among data, then we can predict the future values given the historical values. So the prediction problem can be formulated as a regression problem: give $k$ days' SST values, what are the SST values for the $k+1$ to $k+l$ days? Here $l$ represents the prediction length.

\subsection{Long short-term memory}

To capture the temporal relationship among time series data, we adopt LSTM to do this job. This subsection introduces LSTM briefly.

LSTM was first proposed by Hochreiter in 1997~\cite{Hochreiter1997Long}. It is a specific recurrent neural network  architecture that was designed to model sequences and their long-range dependencies more accurately than conventional RNNs. LSTM can process a sequence of input and output pairs ${(x_i,y_i)}_{i=1}^{n}$. For each pair $(x_i, y_i)$, the LSTM cell takes a new input $x_i$ and the hidden vector $h_{i-1}$ from the last time step, then produces an estimate output $\hat{y_i}$ for the target output $y_i$ given all the previous input sequence $x_1, x_2, \cdots, x_i$ also with a new hidden vector $h_i$ and a new memory vector $m_i$. Fig.~\ref{fig.lstm} shows the structure of a LSTM cell. The whole computation can be defined by a series of equations as follows~\cite{Graves2013Generating}:

\begin{figure}[!hbp]
\centering
\includegraphics[width=2.5in]{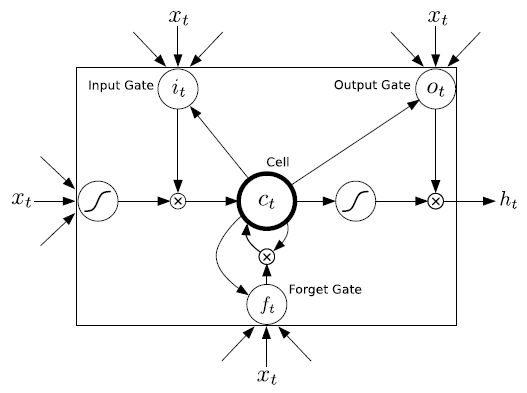}
\caption{Stucture of LSTM cell~\cite{Hochreiter1997Long}} \label{fig.lstm}
\end{figure}

\begin{equation}\label{eq.lstm}
    \begin{split}
        &i = \sigma (W^iH + b^i) \\
        &f = \sigma (W^fH + b^f) \\
        &o = \sigma (W^oH + b^o) \\
        &c = tanh(W^cH + b^c) \\
        &m' = f \odot m + i \odot c \\
        &h' = tanh (o \odot m') \\
    \end{split}
\end{equation}
where $\sigma$ is the sigmoid function, $W^i, W^f, W^o, W^c$ in $\mathbb{R}^{d \times 2d}$ are the recurrent weight matrices, and $b^i, b^f, b^o, b^c$ are the corresponding bias terms. $H$ in $\mathbb{R}^{2d}$ is the concatenation of the new input $x_i$ and the previous hidden vector $h_{i-1}$:

\begin{equation}
 H=\left[   \begin{array}{c} Ix_i \\ h_{i-1} \\ \end{array}     \right]
\end{equation}

The key to LSTM is the cell state, i.e. memory vector $m$ and $m'$ in Equation (\ref{eq.lstm}), which can remember long-term information. The LSTM does have the ability to remove or add information to the cell state, carefully regulated by structures called gates. The gates in Equation (\ref{eq.lstm}) are $i, f, o, c$, representing \textit{input gate}, \textit{forget gate}, \textit{output gate} and a \textit{control gate}. Input gate can decide how much input information enter the current cell. Forget gate can decide how much information be forgotten for the previous memory vector $m_{i-1}$, while the control gate can decide to write new information into the new memory vector $m_i$ modulated by the input gate. Output gate can decide what information will be output from current cell.

Followed the work of~\cite{Kalchbrenner2015Grid}, we also use a whole function $LSTM()$ as shorthand for Equation (\ref{eq.lstm}):
\begin{equation}\label{eq.lstmf}
  (h', m') = LSTM(\left[   \begin{array}{c} Ix_i \\ h_{i-1} \\ \end{array}     \right], m, W)
\end{equation}
where $W$ concatenates the four weight matrices $W^i, W^f, W^o, W^c$.

\subsection{Basic LSTM blocks}

\begin{figure}[!hbp]
\centering
\includegraphics[width=1in]{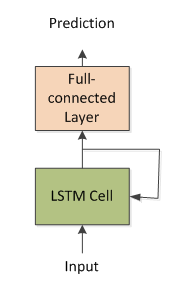}
\caption{Structure of a basic LSTM block} \label{fig.block}
\end{figure}

\begin{figure}[!hbp]
\centering
\includegraphics[width=1in]{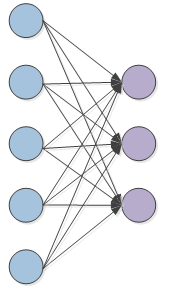}
\caption{A full-connected layer} \label{fig.fc}
\end{figure}

We combine LSTM with full-connected layer to build a basic LSTM block. Fig.~\ref{fig.block} shows the structure of a basic LSTM block. There are two basic neural layers in a block. LSTM layer can capture the temporal relationship, i.e. the regulate variation among the time series SST values. While the output of LSTM layer is a vector i.e. the hidden vector of the last time step, we use a full-connected layer to make a better abstraction and combination for the output vector, and reduce its dimensionality meanwhile map the reduced vector to a final prediction. Fig.~\ref{fig.fc} shows a full-connected layer. The computation can be defined as follows:
\begin{equation}\label{eq.block}
  \begin{split}
    &(h_i, m_i) = LSTM(\left[   \begin{array}{c} Iinput \\ h_{i-1} \\ \end{array}     \right], m, W) \\
    &prediction = \sigma (W^{fc}h_l + b^{fc})  \\
  \end{split}
\end{equation}
where the definition of function $LSTM()$ is as Eqation(\ref{eq.lstmf}), $h_l$ is the hidden vector in the last time step of LSTM, $W^{fc}$ is the weight matrices in full-connection layer, and $b^{fc}$ is the corresponding bias terms.

This kind of block can predict future SST for a single grid, given all the historical SST values of this grid. But it's still not enough. We need to predict SST for an area. So we can use the basic LSTM blocks to construct the whole network.

\subsection{Network architecture}

\begin{figure}[!hbp]
\centering
\includegraphics[width=3in]{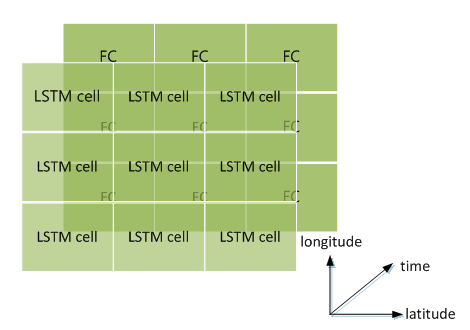}
\caption{Network architecture} \label{fig.network}
\end{figure}

Fig.~\ref{fig.network} shows the architecture of the network. It's like a cuboid: the $x$ axis stands for latitude, the $y$ axis stands for longitude, and the $z$ axis is time direction. Each grid is corresponding to a grid in real data. Actually the grids in the same place along the time axis form a basic block. We omit the connections between layers for clarity.

\section{Experimental Results and Discussions}

\subsection{Study area and data}

We use NOAA High Resolution SST data provided by the NOAA/OAR/ESRL PSD, Boulder, Colorado, USA, from their Web site at http://www.esrl.noaa.gov/psd/~\cite{sstdata}. This data set contains daily values from 1981/09 to 2016/11 (12868 days total), and covers the global ocean from 89.875S to 89.875N, 0.125E to 359.875E, which is 0.25 degree latitude multiplied by 0.25 degree longitude global grid (1440x720).

As we all know, the temperature varies relatively stable in far ocean, while fluctuates more greatly in coastal seas. So we focus on the coastal seas near China to evaluate the proposed method. Bohai sea is the innermost gulf of the Yellow Sea and Korea Bay on the coast of Northeastern and North China. It is approximately 78,000 $km^2$ in area and its proximity to Beijing, the capital of China, makes it one of the busiest seaways in the world~\cite{bohai}. Bohai sea covers from 37.07N to 41N, 117.35E to 121.10E. We take the corresponding subset to the Bohai sea from the dataset mentioned above to form a 16 by 15 grid and contains a total of 12868 daily values, named Bohai SST Dataset.

\subsection{Experimental Setup}

Since we formulate the SST prediction as a sequence prediction problem, i.e. using historical observations to predict the future, we should determine how long the historical observations are to be used to predict the future. Of course the longer the length is, the better the prediction will be, while the more computation it will need. Here we use about 4 times of the prediction length to be the length of the historical observations according to the characteristics of the periodical change of temperature data. In addition, there are still other important values to be determined: the number of layers for LSTM layer $l_r$ and full-connected layer $l_{fc}$, which will determine the whole structure of the network. Also the corresponding number of hidden units denoted by $units\_r$ should be determined together.

According to these aspects mentioned above, we first design a simple but important experiment to determine the critical values for $l_r$ ,$l_{fc}$ and $units\_r$, using the basic LSTM block to predict the SST for a single position. Then we evaluate the proposed method on area SST prediction for Bohai sea.

Once we determine the structure of the network, there are still other critical things to be determined in order to train the network, i.e. the optimization method, the learning rate, the batch size, etc. The traditional optimization method for deep network is stochastic gradient descent (SGD), which is the batch version of gradient descent. The batch method can speed up the convergence of network training. Here we adopt Adagrad optimization method~\cite{Duchi2010Adaptive}, which can adapt the learning rate to the parameters, performing larger updates for infrequent and smaller updates for frequent parameters. Dean et al.~\cite{Dean2012Large} have found that Adagrad greatly improved the robustness of SGD and used it for training large-scale neural networks. We set the initial learning rate as 0.1, and the batch size as 100 in the following experiments.

The division of training set, validation set and test set are as follows. The data from 1981 to 2012.8 (11323 days) is used as training set, the data from 2012.9 to 2012.10 (122 days) is the validation set, and the data from 2013 to 2015 (1095 days) is the test set. We will test for one week (7 days)and one month (30 days) to evaluate the prediction performance. The data of 2016 (328 days) is reserved for another comparison.

Results of another traditional regression model, i.e. support vector regression (SVR), for SST prediction is given for comparison purpose. We run the experiments under the environment of Intel(R) Core(TM)2 Quad CPU Q9550 @2.83GHz, 6G RAM, Ubuntu 16.10 64 bits operating system, and Python 2.7. The proposed network is implemented by TensorFlow 0.11~\cite{tensorflow2015-whitepaper}. SVR is implemented by Scikit-learn~\cite{scikit-learn}.

The performance evaluation of SST prediction is a fundamental issue. In this letter, we use root of mean squared error (RMSE), one of the most commonly used measurement as the evaluation metric to measure the effectiveness of different methods. In addition, we define a metric to evaluate the prediction accuracy as follows:
\begin{equation}
 ACC = \frac{1}{n}\sum_{i=1}^{n}\frac{|prediction-true~value|}{true~value}
\end{equation}
where $n$ is the prediction length.

RMSE is the smaller the better, while ACC is the opposite. Here RMSE and ACC can be regarded as absolute error and relative error. And for area prediction, we use the area average RMSE and area average ACC.

\subsection{Determination of parameters}

We randomly choose 5 positions in Bohai Sea denoted as $p_1, p_2,...,p_5$ to predict 7 days' SST values with lead time of one month (30 days). Firstly, we fix  $l_r$ and $l_{fc}$ as 1, $units\_fc$ as 7, and choose a proper value for $units\_r$ from \{3,4,5,6\}. Table~\ref{tb.1} shows the results on five positions with different values of $nunits\_r$. The boldface items in the table represent the best performance, i.e. the smallest RMSE and the largest ACC.
It can be seen from the results that the most best performance occurs when $units\_r=6$.

In this experiment, the best performance occurs when $units\_r=6$ in three positions $p_1, p_2$ and $p_3$, while occurs when $units\_r=5$ at $p_5$ and $units\_r=7$ at $p_4$. But we can see that the difference of RMSE and ACC is not too large. So in the following experiments, we set $units\_r6$ as 6.
\begin{table}
    \centering
    \caption{Prediction Results (RMSE\&ACC) on Five Positions with Different $units\_r$s.}
    \label{tb.1}
    \renewcommand\arraystretch{1.2}
    \begin{tabular}{c|c|c c c c c } \hline \hline
     $units\_r$  & metrics & $p_1$ & $p_2$  & $p_3$  & $p_4$  & $p_5$   \\ \hline
     \multirow{2}{*}{3} &RMSE  &0.5757 &0.5907 &3.3174 &0.8656 &5.6695  \\
                        &ACC  &0.9824 &0.9825 &0.9276 &0.9725 &0.8593 \\ \hline
     \multirow{2}{*}{4} &RMSE   &0.6026 &0.5412 &0.8191 &0.8156 &0.8222  \\
                        &ACC &0.9820 &0.9838 &0.9718 &0.9737 &0.9718 \\ \hline
     \multirow{2}{*}{5} &RMSE  &0.5866 &0.5463 &0.8445 &0.8581 &\textbf{0.7441} \\
                        &ACC &0.9818 &0.9834 &0.9711 &0.9729 &\textbf{0.9742} \\ \hline
     \multirow{2}{*}{6} &RMSE &\textbf{0.5649} &\textbf{0.5254} &\textbf{0.7663} &0.8125 &0.8176 \\
                        &ACC &\textbf{0.9829} &\textbf{0.9843} &\textbf{0.9730} &0.9738 &0.9721 \\ \hline
     \multirow{2}{*}{7} &RMSE &0.5820 &0.5302 &0.7790 &\textbf{0.7816} &0.7470 \\
                        &ACC  &0.9825 &0.9841 &0.9728 &\textbf{0.9749} &0.9741 \\ \hline
                        \hline
  \end{tabular}
\end{table}

Then, we also use the SST sequences from the same five positions to choose a proper value for $l_r$ from \{1,2,3\}. The other two parameters are set by $unit\_r=6$, and $l_{fc}=1$. Table~\ref{tb.3} shows the results on five positions with different values of $l_r$. The boldface items in the table represent the best performance. It can be seen from the results that the best performance occurs when $l_r=1$. The reason may due to the increasing weights numbers with increasing recurrent LSTM layers, which our data is insufficient to train so many weights. Actually, experiences show that the recurrent LSTM layer is not the more the better. And during the experiments we find that more LSTM layers are more likely to get unstable results. So in the following experiments, we set $l_r$ as 1.
\begin{table}
    \centering
    \caption{Prediction Results (RMSE\&ACC) on Five Positions with Different $l_r$s.}
    \label{tb.3}
    \renewcommand\arraystretch{1.2}
    \begin{tabular}{c|c|c c c c c } \hline \hline
     $l_r$  & metrics & $p_1$ & $p_2$  & $p_3$  & $p_4$  & $p_5$   \\ \hline
     \multirow{2}{*}{1} &RMSE &\textbf{0.5649} &\textbf{0.5254} &\textbf{0.7663} &\textbf{0.8125} &\textbf{0.8176} \\
                        &ACC &\textbf{0.9829} &\textbf{0.9843} &\textbf{0.9730} &\textbf{0.9738} &\textbf{0.9721} \\ \hline
     \multirow{2}{*}{2} & RMSE &3.0357 &0.5773 &3.3091 &0.8289 &4.0466  \\
                              &ACC &0.9442 &0.9826 &0.9296 &0.9736 &0.9177 \\ \hline
     \multirow{2}{*}{3} &RMSE &3.0371 &0.5711 &0.7991 &0.8442 &4.0451 \\
                              &ACC &0.9443 &0.9832 &0.9721 &0.9730 &0.9163 \\ \hline \hline
  \end{tabular}
\end{table}

Lastly, we still use the SST sequences from the same five positions to choose a proper value for $l_{fc}$ from \{1,2\}. Though the number of the hidden units of the full-connected layer is tricky. Table~\ref{tb.2} shows the results with different $l_{fc}$s. The numbers in the square brackets stand for the number of the hidden units. The boldface items in the table represent the best performance. It can be seen from the results that it achieve the most best performance when $l_{fc} = 1$. The reason may be the same: more layers means more weights to be trained and more computation it needs. So in the following experiments, we set $l_{fc}$ as 1, and the number of its hidden units is set the same as the prediction length.

\begin{table}
    \centering
    \caption{Prediction Results (RMSE\&ACC) on Five Positions with Different $k$s.}
    \label{tb.2}
    \renewcommand\arraystretch{1.2}
    \begin{tabular}{c|c|c c c c c } \hline \hline
     $l_{fc}$  & metrics & $p_1$ & $p_2$  & $p_3$  & $p_4$  & $p_5$   \\ \hline
     \multirow{2}{*}{1[7]} &RMSE &0.5649 &\textbf{0.5254} &\textbf{0.7663} &\textbf{0.8125} &\textbf{0.7376} \\
                        &ACC &0.9829 &\textbf{0.9843} &\textbf{0.9730} &\textbf{0.9738} &\textbf{0.9789} \\ \hline
     \multirow{2}{*}{2[7,7]} &RMSE &\textbf{0.5533} &0.5266 &0.7805 &3.1091 &6.9153  \\
                        &ACC &\textbf{0.9832} &0.9842 &0.9730 &0.9357 &0.8044 \\ \hline
     \multirow{2}{*}{2[10,7]} &RMSE &0.5794 &0.5298 &3.3422 &6.0412 &5.6626 \\
                        &ACC &0.9823 &0.9840 &0.9265 &0.8235 &0.8617 \\ \hline
     \multirow{2}{*}{2[15,7]} &RMSE &3.0349 &2.6857 &0.7856 &3.1001 &0.7430 \\
                        &ACC &0.9454 &0.9510 &0.9645 &0.9373 &0.9742 \\ \hline
\hline
  \end{tabular}
\end{table}

\subsection{Results and Analysis}

We use Bohai SST data set to do this experiment, and compare the proposed method to a classical regression methods SVR~\cite{Drucker1996Support}. The setting is as follows. For LSTM network, we set $k=10,15,30,120 for l=1,3,7,30 represently, and l_r=1, l_{fc}=1$. For SVR, we use the RBF kernel and set the kernel width $\sigma=1.6$ which is chosen by cross validation on validation set.

\begin{figure}[!hbp]
\centering
\includegraphics[width=3.5in]{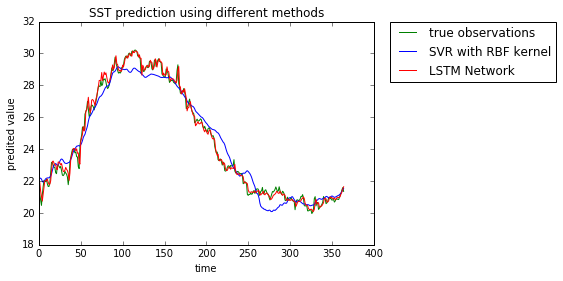}
\caption{SST Onr Month Prediction at One Position Using Different Methods} \label{fig.result}
\end{figure}

Table~\ref{tb.4} shows the results. The boldface items in the table represent the best performance, i.e. the smallest area average RMSE and the largest area average ACC. It can be seen from the results that the LSTM network achieve the best prediction performance. And Fig.\ref{fig.result} shows the SST prediction at one position using two different methods. In order to see the results clearly, we only show the prediction results for one year. Green line represents the true value. Red line represents the prediction results of the LSTM network, and blue line represents the prediction results of SVR with RBF kernel.

\begin{table*}
    \newcommand{\tabincell}[2]{\begin{tabular}{@{}#1@{}}#2\end{tabular}}
    \centering
    \caption{Prediction Results (Area Average RMSE \& ACC) on Bohai Sea Data Set.}
    \label{tb.4}
    \renewcommand\arraystretch{1.2}
    \begin{tabular}{c|c|cccc } \hline \hline

      \multirow{2}{*}{Methods} & \multirow{2}{*}{Metrics} & \multicolumn{4}{c}{Prediction Length} \\ \cline{3-6}
       \multicolumn{1}{c|}{} & \multicolumn{1}{c|}{} & \tabincell{c}{1\\(one day)} & \tabincell{c}{3\\(three days)}& \tabincell{c}{7\\(one week)}   & \tabincell{c}{30\\(one month)}  \\ \hline

     \multirow{2}{*}{SVR} &RMSE  &0.3998  &0.6158  &0.8388 &1.2477 \\
                          &ACC  &0.9872  &0.9802    &0.9728 &0.9593 \\  \hline
     \multirow{2}{*}{LSTM network} &RMSE  &\textbf{0.0767}  &\textbf{0.1775}    &\textbf{0.6540} &\textbf{1.1363} \\
                                   &ACC  &\textbf{0.9923}  &\textbf{0.9878}    &\textbf{0.9795} &\textbf{0.9690} \\ \hline\hline
  \end{tabular}
\end{table*}

\subsection{Online model update}

In this experiment, we want to show the online characteristics of the proposed method. We have SST values of 328 days in 2016. We call the model trained above \textit{original model}, and use this model to predict the SST values of 2016. Based on the original model, we continue to train the model adding three years' SST observations data of 2013, 2014 and 2015, and get a new model called \textit{updated model}. Table~\ref{tb.5} shows the results of SST prediction for 2016 using these two different models. The updated model performs the best as expected.

This shows a kind of online characteristics of the proposed method: performing prediction, collecting true observations, feeding the true observations back into the model to update it, and going on. While other regression models like SVR don't have such characteristics: when collecting new observations, the model could only be retrained from scratch, which will waste additional computing resources.

\begin{table}
    \newcommand{\tabincell}[2]{\begin{tabular}{@{}#1@{}}#2\end{tabular}}
    \centering
    \caption{Prediction Results(Area Average RMSE \& ACC) on Bohai Sea Data Set in 2016.}
    \label{tb.5}
    \renewcommand\arraystretch{1.2}
    \begin{tabular}{c|c|cccc } \hline \hline

      \multirow{2}{*}{Model} & \multirow{2}{*}{Metrics} & \multicolumn{4}{c}{Prediction of 2016} \\ \cline{3-6}
       \multicolumn{1}{c|}{} & \multicolumn{1}{c|}{} & \tabincell{c}{1} & \tabincell{c}{3}& \tabincell{c}{7}   & \tabincell{c}{30}  \\ \hline

     \multirow{2}{*}{original} &RMSE &0.1346 &0.2145 &0.6891 &1.1521 \\
                                     &ACC &0.9812 &0.9887 &0.9711 &0.9606 \\ \hline
     \multirow{2}{*}{updated} &RMSE &\textbf{0.0899} &\textbf{0.1843} &\textbf{0.5825} &\textbf{1.0123} \\
                                    &ACC &\textbf{0.9905} &\textbf{0.9804} &\textbf{0.9798} &\textbf{0.9701}     \\ \hline\hline
  \end{tabular}
\end{table}

\section{Conclusion}

In this letter, we formulate the prediction of SST as a time series regression problem, and propose a LSTM based network to model the temporal relationship of SST to predict the future value. This is the first time, to our knowledge, to use recurrent neural network to solve the prediction problem of SST. The proposed network utilizes LSTM layer to model the time series data, and full-connected layer to map the output of LSTM layer to a final prediction. We explore the optimal setting of this architecture by experiments and report the prediction performance of coastal seas of China to confirm the effectiveness of the proposed method. In addition, we also show the online update characteristics of the proposed method.

And furthermore, the proposed network  is independent of the resolution of data. If a high resolution prediction is wanted, all that is needed is to provide a high resolution training data to the network. Once we get the predicted SST values in the future, it can be used in many applications including ocean front prediction, abnormal event prediction, etc.


\end{document}